\documentclass[10pt,twocolumn,letterpaper]{article}

\usepackage{iccv}
\usepackage{times}
\usepackage{epsfig}
\usepackage{graphicx}
\usepackage{amsmath}
\usepackage{amssymb}
\usepackage{xcolor}
\usepackage{booktabs}
\usepackage{algorithm}
\usepackage{algorithmic}

\definecolor{myorange}{RGB}{237, 125, 49}
\definecolor{myblue}{RGB}{0, 176, 240}
\definecolor{mypurple}{RGB}{112, 48, 160}
\definecolor{mygreen}{RGB}{84, 130, 53}

\usepackage{multirow}
\usepackage[pagebackref=true,breaklinks=true,letterpaper=true,colorlinks,bookmarks=false]{hyperref}

\iccvfinalcopy 

\def\iccvPaperID{2267} 

\ificcvfinal\fi
\begin{document}

\title{Few-Shot Learning with Global Class Representations}

\author{Tiange Luo$^{1}$\thanks{Equal contribution.}~~~~~Aoxue Li$^{1}$\footnotemark[1]~~~~~Tao Xiang$^{2}$~~~~~Weiran Huang$^{3}$~~~~Liwei Wang$^{1}$\\
  $^1$School of EECS, Peking University, Beijing, China\\
  $^2$Department of Electrical and Electronic Engineering, University of Surrey, UK\\
  $^3$Huawei Noah's Ark Lab, Beijing, China\\
  {\tt\small \{lax,luotg,wanglw\}@pku.edu.cn, t.xiang@surrey.ac.uk,huang.inbox@outlook.com}
}

\maketitle
\thispagestyle{empty}

\begin{abstract}

In this paper, we propose to tackle the challenging few-shot learning (FSL) problem by learning global class representations using both base and novel class training samples. In each training episode, an episodic class mean computed from a support set is registered with the global representation via a registration module. This produces a registered global class representation for computing the classification loss using a query set. Though following a similar episodic training pipeline as existing meta learning based approaches, our method differs significantly in that novel class training samples are involved in the training from the beginning. To compensate for the lack of novel class training samples, an effective sample synthesis strategy is developed to avoid overfitting. Importantly, by joint base-novel class training, our approach can be easily extended to a more practical yet challenging FSL setting, i.e., generalized FSL, where the label space of test data is extended to both base and novel classes. Extensive experiments show that our approach is effective for both of the two FSL settings. 


\end{abstract}

\section{Introduction}

Deep learning has achieved great success in various recognition tasks \cite{Szegedy2017aaai,szegedy2015cvpr}. However, with a large number of parameters, deep neural networks need large amounts of labeled data from each class for model training. This severely limits their scalability -- for many rare classes, collecting a large number of training samples is infeasible or even impossible.  In contrast, humans can easily recognize a new object class after only seeing it once. Inspired by the few-shot learning ability of humans, there has been increasing interest recently on few-shot learning (FSL) \cite{Edwards2017iclr,Kaiser2017iclr,Ravi2017iclr,Dai2017nips,Rezende2016icml,rahman2017unified,li2019large, Garcia2017arxiv,Ren2018arxiv, gidaris2018dynamic}. In the FSL problem, we are provided with a set of base classes with ample training samples per class, and a set of novel classes with only a few labeled samples (shots) per class. FSL aims to learn a classifier for the novel classes with few shots by transferring knowledge from the based classes. 

\begin{figure}[t]
\begin{center}
\includegraphics[width=0.9\columnwidth]{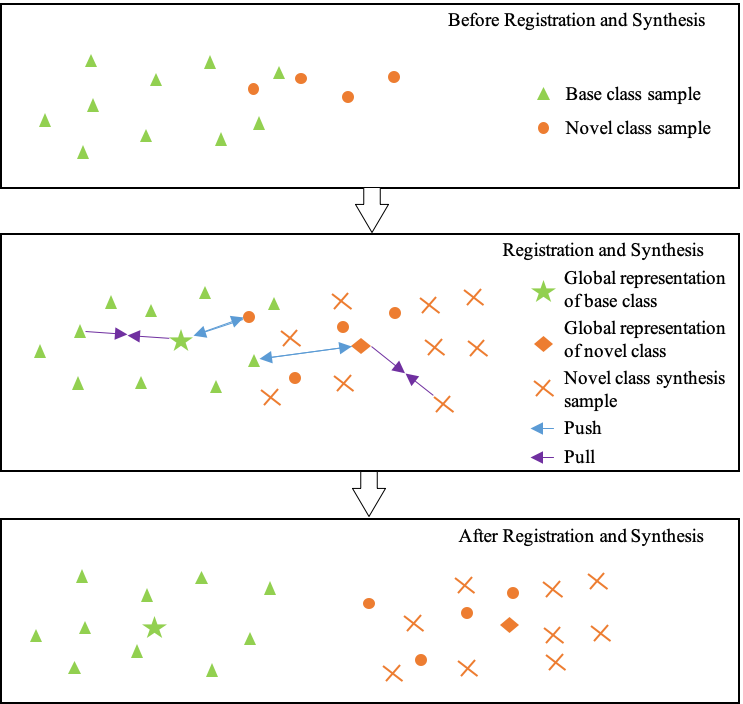}
\end{center}\vspace{-0.2cm}
\caption{An illustration of our approach. \textbf{The first block} shows a base class and a novel class in an embedding space. The base class contains sufficient labeled data while the novel class has only a few labeled data. The two classes have intersections, and we aim to learn global representations for each class which are used for recognizing test data. \textbf{The second block} illustrates the two key components of the proposed model. First, we generate new samples (orange cross) to increase intra-class variance for novel classes. Second, a registration module is proposed to encourage sample to `pull' its global representation to itself and `push' other global representations away. Similarly, global representations would influence the sample. \textbf{The last block} shows the results after learning global representations jointly using both base and novel class samples. The two classes become more separable, and the global representations are more distinguishable.}
\vspace{-0.5cm}
\label{fig1}
\end{figure}


Most existing FSL approaches are based on meta learning. During a meta learning phase, the base classes are sampled to simulate the few-shot learning condition for the novel classes.  Transferable knowledge is then learned from the source classes in the form of good initial conditions \cite{Finn2017icml}, embeddings \cite{vinyals2016bnips,Snell2017nips} or optimization strategies \cite{Ravi2017iclr}. After the meta learning phase, the target few-shot learning problem is solved by fine-tuning \cite{Finn2017icml} with the learned optimization strategy \cite{Ravi2017iclr} or computed in a feed-forward pass \cite{Snell2017nips,vinyals2016bnips,Santoro2016arxiv} without updating network weights. However, there is a fundamental limitation in these meta learning based approaches: the model (initial condition, embedding or optimization strategy) are mostly learned with the source data only. This offers no guarantee for the model to generalize well on the target data, even after the fine-tuning step.

In this paper, we propose a novel approach to FSL by representing each class, base or novel, as a single point in an embedding space. Since the representation is learned jointly using both base and novel class training samples, it is called a global representation. We argue that only by involving the novel class data at the very beginning of the model training, we can ensure that the learned FSL model is suited for the novel classes. 

A critical obstacle for learning such a global class representation is the imbalanced training sample numbers across the base and novel classes. We overcome this problem by two means. First, we use sample synthesis to increase the intra-class variation for novel classes. By randomly sampling data points from a subspace of samples in the same classes, our synthesis strategy can effectively increase intra-class variance (see orange crosses in Figure \ref{fig1}). Second, we introduce episodic training to balance the base and novel class samples. In each training episode, an episodic class mean computed from a support set is registered with the global representation via a registration module. This produces a registered global class representation for computing the classification loss using a query set. By learning to compare each data against \emph{all} global class representations, our registration module forces each data to `pull' the global representation of its class toward itself and `push' other global representations away in the embedding space (see blue arrows in Figure \ref{fig1}). After the training, the learned global representations are used for recognizing test data.

Since the base and novel classes are involved simultaneously in every step of the training process, the learned global representations are naturally able to distinguish both the base and novel classes. This means that our approach can be easily extended to a more realistic yet more challenging FSL setting (i.e., Generalized FSL) where the label space of test data covers both base and novel classes. This is as opposed to the standard setting where the test data contain novel class samples only. Under this setting, there is no way that one can tell whether the learned global representations have been biased towards to the base classes.  

Our main contributions are as follows: 1) We propose a novel FSL approach that recognizes novel class data by learning global representations using both base and novel classes training samples.  2) Our approach can be easily extended to the more realistic generalized FSL setting.  Extensive experiments on two FSL benchmarks show that the proposed approach is effective under both the standard and generalized settings. Importantly, the improvement is even bigger under the generalized FSL setting.

\section{Related Work}

Recently, few-shot object recognition has become topical. With the success of deep learning-based approaches in the data-rich many-shot setting \cite{Russakovsky2015ImageNet,he2016cvpr}, there has been a surge of interest in generalizing such deep learning approaches to the few-shot learning setting, so that visual recognition can truly scale to a large number of classes (e.g., millions). Most of the recent deep learning based approaches use a meta-learning or learning-to-learn strategy. With meta learning,  these models extract transferrable knowledge from a set of auxiliary tasks via episodic training, which then helps them to solve the target few-shot classifier training for the target novel classes. 

Existing meta-learning based FSL approaches can be grouped into three categories: 1) The first category address the FSL problem by ``learning to fine-tune''. These approaches aim to learn good model initialization (i.e., parameters of a network) so that the classifiers for novel classes can be learned with a limited number of labeled examples and a small number of gradient update steps \cite{Finn2017icml,Rusu2019iclr,Nichol2018arxiv}. 2) The second category of models tackles the FSL by ``learning a good metric to compare''. The intuition is that if a model can determine the similarity of two images, it can classify an unseen input image with the labeled instances \cite{Snell2017nips,Sung2018cvpr}. To learn an effective comparison model, these methods make their prediction conditioned on distances to few labeled instances during the training process \cite{Bertinetto2019iclr,vinyals2016bnips,Koch2015icml,Bertinetto2016nips}. These instances are sampled from base classes designed to simulate the test scenario where only a few shots from the novel classes are available.  3) The third category of models deals with the FSL problem by ``learning an optimizer''. These models attempt to modify the classical gradient-based optimization (e.g., stochastic gradient descent) to fit into the meta-learning scenario \cite{Ravi2017iclr,Munkhdalai2017icml}. Despite their dominance in the recent literature, a fundamental problem remains for these meta-learning-based models: in the training stage, only the base classes samples are involved, making them vulnerable to overfitting to the base classes. 

The most related approach to ours is the prototypical networks \cite{Snell2017nips}, which aims to learn a class representation, or prototype by feeding the feature mean of a few shots of the class to a fully-connected layer. Compared with \cite{Snell2017nips}, our model has two vital differences: (1) We learn global class representation rather than episodic one as in \cite{Snell2017nips}. (2) Both base and novel class training samples are used to jointly learn the representation. This ensures that a class representation can be learned with a global consistency rather than a local one. Also related is the recent feature hallucination based approach \cite{Wang2018cvprlsl}. These are orthogonal to ours -- we can employ any one of them in our model for synthesizing novel class samples to tackle the class imbalance problem. 


\section{Methodology}

The key idea of our model is joint class representation learning using both base and novel class training samples. To overcome the class imbalance problem, we employ representation registration and novel class sample synthesis. In this section, we first introduce these two key modules. After that, we describe how to integrate the two modules into our FSL framework to recognize unlabeled data from novel classes. Finally, we extend the proposed approach to the generalized FSL setting, where unlabeled data are from both base and novel classes. 

\subsection{Registration Module}
\label{sec:reg}
Suppose we have a set of classes $C_{total}=\{c_1,...,c_N\}$, where $N$ denotes the total number of classes. These include both base and novel classes. We are given a training set $D_{train}$ whose label space is $C_{total}$ (i.e., both base and novel classes are used for training) and a test set $D_{test}$.  Our registration module compares the training samples against global representations of all training classes and selects the corresponding global representation. A registration loss is defined to jointly optimize global representations and the registration module.

Concretely, first, a sample $x_i$ in the training set is fed into a feature extractor $F$ to obtain its visual feature, denoted as $f_i = F(x_i)$. Then, visual feature of this sample and all global class representations $G = \{g_{c_j}, c_j\in C_{total}\}$ are fed into registration module $R$. For each visual feature $f_i$ , the registration module $R$ produces a vector $V_i=[v_i^{c_1},....,v_i^{c_N}]^T$, where the $j$-th element is the similarity score between $f_i$ and the global representation $g_{c_j}$ for the class $c_j$. In this paper, we compute similarity score in an embedding space.
\begin{equation}
\small\label{reg1}
\begin{split}
&v_i^{c_j}= \frac{\exp{\left(d_i^{c_j}\right)}}{\sum_{c_j\in C_{total}}\exp{\left(d_i^{c_j}\right)}} \\
&d_i^{c_j} = -{\|\theta(f_i)-\phi(g_{c_j})\|}_2
\end{split}
\end{equation} 
where $\theta(\cdot)$ and $\phi(\cdot)$ are embeddings for visual feature of samples and global class representations, respectively. 

Therefore, we define a registration loss $L_{reg}$ for sample $x_i$ (with its label $y_i$) to make the sample nearest to its global class representation in the embedding space, where $CE$ denotes a cross entropy loss.
\begin{equation}
\small
\label{reg2}
\begin{split}
&\mathcal{L}_{reg}=CE(y_i, V_i)
\end{split}
\end{equation} 

By comparing sample against global representations of all classes in $C_{total}$ in the embedding space, our registration module makes each global representations close to samples within its class, and away from the extra-class samples. Note that both the representation and the feature extraction network are end-to-end trainable and jointly optimized. Specifically, with well trained global class representations, the feature extractor is optimized to cluster samples around these class representations; given feature extractor, each global class representation is optimized to be closer to samples in its class and away from other representations.

When integrating the registration module into the FSL framework (to be detailed later), we feed labeled data into the registration module to select global representations for classifying query images. The loss of classifying query images will also optimize global representations, together with the registration loss.

\begin{figure*}[t]
\begin{center}
\includegraphics[width=0.98\textwidth]{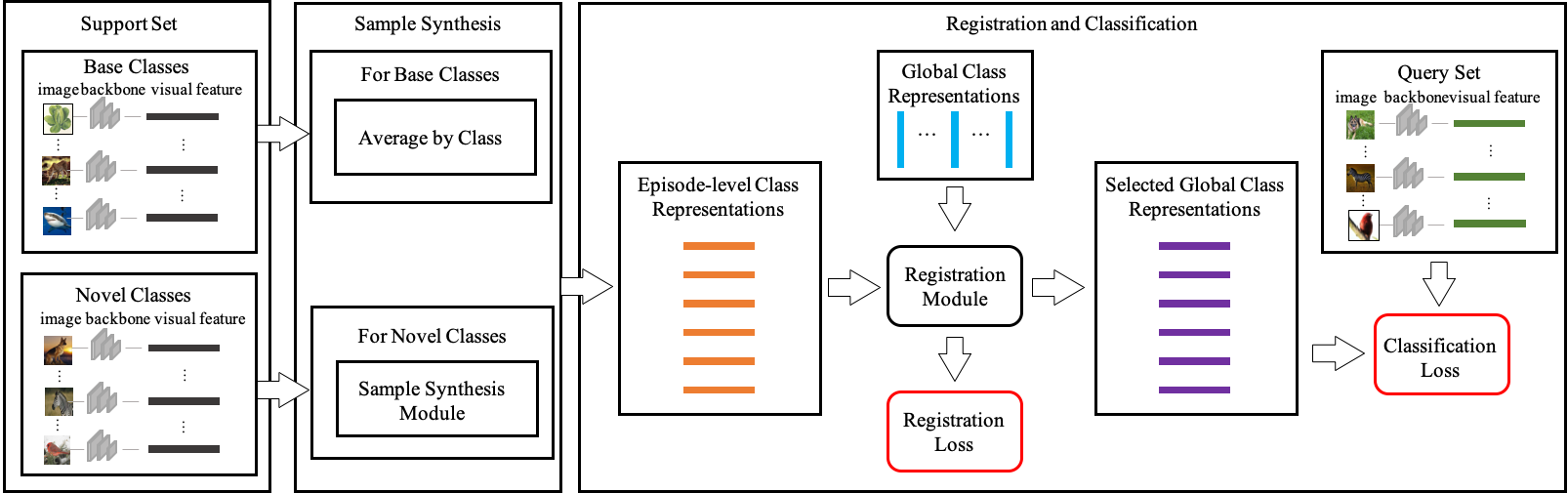}
\end{center}
\vspace{-0.05in}
\caption{Overview of the whole framework. First, we propose a sample synthesis method to synthesize \textcolor{myorange}{episodic representation} for each class in the support set. Second, the registration module is leveraged to select \textcolor{myblue}{global representation} according to their episodic representation, and the \textcolor{mypurple}{selected global representations} are then used to classify \textcolor{mygreen}{query images}. The classification loss and registration loss are used to jointly optimize the global representations, the registration module, and the feature extractor. (Best viewed in color)}
\label{fig2}
\vspace{-0.1in}
\end{figure*}

\subsection{Sample Synthesis Module}

\label{sec:syn}
To address the class imbalance issue caused by the limited data (few shots) in the novel classes, we propose a sample synthesis strategy to synthesize samples for novel classes denoted as $C_{novel}$. In this paper, we synthesize samples by two steps: 1) we generate new samples with original samples. 2) we synthesize a new sample by using all samples obtained by the first step. 

Specifically, we first generate new samples with original samples by using random cropping, random flipping and data hallucination \cite{Wang2018cvprlsl}. These methods take a single example of a class as input and produce variants of this example. Using the three methods on the original few training samples, we will obtain a total of $k_t$ samples for every novel class. After that, we further synthesize new samples from the $k_t$ samples per class. In particular, for a novel class $c_j$, we first randomly select $k_r$ samples out of the $k_t$ samples. Then, we synthesize a new sample by randomly selecting a data point from the subspace spanned by the $k_r$ visual features of the samples $\{f_1,..,f_{k_r}\}$. Concretely, we sample $k_r$ values $\{\nu_1, ..., \nu_{k_r}\}$ from a uniform distribution ranging from 0 to 1. Then, we weight sum these visual features by using random numbers as weights. A new sample $r_{c_j}$ for the novel class $c_j$ is defined in Equation \ref{synthesis}. With the proposed strategy, the intra-class variation thus increases and the limited data issue is alleviated (as validated in Figure \ref{ab_study}). 
\begin{equation}
\label{synthesis}
\begin{split}
\small
&r_{c_j}= \sum_{i=1}^{k_r}\frac{\nu_i}{\sum_j\nu_j} f_i, \text{where} y_i=c_j\\
&\hat{k}_r\sim\mathcal{U}(0,k_t), k_r =\lceil\hat{k}_r\rceil,\nu_i\sim\mathcal{U}(0,1),  \\
\end{split}
\end{equation}
where $r_{c_j}$ denotes the synthesized sample for novel class $c_j$. $\mathcal{U}(a,b)$ denotes a uniform distribution ranging from $a$ to $b$.

\subsection{Few Shot Learning By Registration}
\label{sec:FSL}

Now we can describe our full FSL framework. In FSL, the class set $C_{total}$ consists of two disjoint sets: a set of base classes $C_{base}$ and a set of novel classes $C_{novel}$. In the training set, each base class has sufficient labeled data, while each novel class is given only $n_{few}(n_{few}\leq5)$ labeled samples. In the test set, samples are from classes in $C_{novel}$ under the standard FSL setting. We first obtain an initial class representation for each class in  $C_{total}$  by simply averaging visual features of all samples from the class. Our model aims to learn the global class representation for each novel class given the simple initialization.

To alleviate the severe data imbalance issue in the training set, apart from the data synthesis strategy described earlier, we define an episodic learning strategy commonly adopted by many existing meta learning models \cite{Snell2017nips}. In each training iteration, an episode/mini-batch is obtained by the following three steps: 1) We first randomly select $n_{train}$ classes from the whole class set $C_{total}$ to form a training episode class set $C_{train}$; 2) $n_s$ samples of each of the classes in $C_{train}$ are randomly selected from the training set to form a support set $S=\{(x_i,y_i), i=1,...,n_s\times n_{train}\}$. 3) We select $n_q$ samples of each of the classes in $C_{train}$ from the training set to form a query set $Q=\{(x_k,y_k), k=1,...,n_q\times n_{train}\} $. Note that, each novel class has only $n_{few}$ labelled samples in training set, where $n_{few}$ is usually smaller than $n_s+n_q$. Therefore, we first augment the $n_{few}$ original samples to $n_s+n_q$ samples by using the synthesis method proposed in Section \ref{sec:syn}, then split them into $n_s$ samples and $n_q$ samples, and put them into the support set and the query set, respectively. In each testing iteration, an episode is different from a training episode in three parts: 1) $C_{test}$ consists of $n_{test}$ randomly selected from novel classes only. 2) We use the labeled images of the $n_{test}$ novel classes in the training set as the support set (i.e., the few shots). 3) The query set is selected from $D_{test}$ instead of $D_{train}$. If there is $n_{few}$ labeled samples per novel class in the training set, the FSL problem is called $n_{few}$-shot FSL. If a model predicts the label of a test image from $n_{test}$ candidate classes during the test stage, the FSL problem is called $n_{test}$-way FSL.

With the episodic learning strategy, we first integrate the proposed sample synthesis module into the FSL framework. Specifically, in each training iteration, the images $\{x_i, i=1,...,n_s\times n_{train}\}$ in the support set $S$ are first fed into the trainable feature extractor $F$  to obtain their visual features $\{f_i = F(x_i), i=1,...,n_s\times n_{train}\}$. Second, we construct a episodic representation for each class in the support set, denoted as $\{r_{c_i}, c_i \in C_{train}\}$.
This episodic representation $r_{c_i}$ integrates information of class $c_i$ in the support set $S$ for the current mini-batch; it is thus a local class representation rather than global.  For base classes, we average the visual features in the same class to obtain the episodic class representations, similar to that in Prototypical Nets \cite{Snell2017nips}. For novel classes, we exploit the synthesis strategy proposed in Section \ref{sec:syn} to synthesize a new sample for each class, with visual features from this class in the support set as input (see 'Sample Synthesis' in the Figure \ref{fig2}). Such episodic novel class representations are more diverse than original labeled samples. 

\begin{algorithm*}[t]
\caption{Training episode loss computation.}
\label{alg:fsl}
\begin{algorithmic}
\STATE {\bfseries Input:} Whole class set $C_{total}$, base class set $C_{base}$, novel class set $C_{novel}$, training set $D_{train}$, test set $D_{test}$, feature extractor $F$, registration module $R$ and global class representations $G=\{g_{c_j}, c_j\in C_{total}\}$.\\
\STATE {\bfseries Output:} The loss for a randomly generated training episode.\\
\STATE  1. Randomly sample $n_{train}$ classes from $C_{total}$ to form $C_{train}$; \qquad\qquad\qquad\qquad\qquad\qquad \\
\STATE  2. Randomly sample $n_s$ images per class in $C_{train}$ to form a support set $S=\{(x_i,y_i), i=1,..,n_s\times n_{train}\}$; \\
\STATE  3. Randomly sample $n_q$ images per class in $C_{train}$ to form a query set $Q=\{(x_j,y_j), j=1,..,n_q\times n_{train})\}$; \\
\STATE  4. Compute visual features of images in $S$ by using the feature extractor $F$, and obtain visual features $\{f_i = F(x_i), i=1,...,n_q\times n_{train}\}$;
\STATE  5. Construct episodic representations $\{r_{c_i}, c_i\in C_{train}\}$ by using the features within their own classes and the sample synthesis module.
\STATE  6. Compute the similarity score vector $V_i=[v_i^{c_1},....,v_i^{c_N}]^T$ between each episodic representation $r_{c_i}$ and all global class representations $G=\{g_{c_j}, c_j\in C_{total}\}$ according to Equation \ref{reg_loss};
\STATE  7. Compute the registration loss according to Equation \ref{reg_loss};
\STATE  8. Select corresponding global class representation $\{\xi_i, i=1,..,n_{train}\}$ by using $\xi_i = V_i G$;
\STATE  9. Compute the classification loss of query images according to Equation \ref{fsl_loss};
\STATE 10. Compute the total loss according to Equation \ref{total_loss}.
\end{algorithmic}
\end{algorithm*}


Then, the registration module is integrated into our FSL framework to select global representation according to its episodic representation. The selected global representations are then used to classify query images. Specifically, we feed episodic representations of classes in the support set $\{r_{c_j}, c_j\in C_{train}\}$ and all global class representations $G = \{g_{c_j},c_j\in C_{total}\}$  into the registration module $R$ to compute similarity score between each episodic class representation and all global class representations according to Equation \ref{reg1}. The similarity score will be used to select global class representations for query images. To make the global class representations more separable, our registration module defines a registration loss to impose the similarity score of its to be bigger than those of other global representations. The more separable global class representations can enhance the ability to recognize unlabeled images.  

According to Equation \ref{reg2}, the registration loss of one class episodic representation $r_{c_i}$ is formulated as follows:
\begin{equation}
\small
\label{reg_loss}
\begin{split}
&\mathcal{L}_{reg}(r_{c_i})=CE(c_i, V_i), \\
&v_i^{c_j}= \frac{\exp{\left(d_i^{c_j}\right)}}{\sum_{c_j\in C_{total}}\exp{\left(d_i^{c_j}\right)}} \\
&d_i^{c_j} = -{\|\theta(r_{c_i})-\phi(g_{c_j})\|}_2
\end{split}
\end{equation} 
where $V_i=[v_i^{c_1},....,v_i^{c_N}]^T$ denotes the similarity scores between the episodic representation $r_{c_i}$ for class $c_i$ in $C_{train}$ and all global class representations $\{g_{c_j}, c_j \in C_{total}\}$. 

Finally, with the similarity score obtained by the third step, we select a global class representation for each class in $C_{train}$ as its class representation and recognize query images by performing the nearest neighbor search using the selected global class representations as references. However, it is nondifferentiable when the selection is an argmax operation. Therefore, we select the class representation in a soft manner: taking the probability distribution $V_i$ as the weight,
we estimate class representations of the $i$-th class in $C_{train}$ (denoted as $\xi_i$) as a weighted sum of global representations of all classes. That is,  $\xi_i = V_i G$. In our experiments (see Section \ref{sec: mini}), the soft manner shows almost the same performance as the argmax operation. Now we obtain the corresponding global class representation set $\{\xi_i, i=1, ..., n_{train}\}$ for classes in $C_{train}$. The classification loss of query samples $\{x_k, y_k\} \in Q$ is formulated as follows:
\begin{equation}
\label{fsl_loss}
\small
\begin{split}
&\mathcal{L}_{fsl}(x_k)=CE(y_k, W_k), \\
&w_k^i = \frac{\exp{\left(d_k^i\right)}}{\sum_{i\in C_{train}}\exp{\left(d_k^i\right)}} \\
&d_k^i = -{\|F(x_k)-\xi_i\|}_2
\end{split}
\end{equation} 
where $F$ denotes the feature extractor, and $W_k=[w_k^1,....,w_k^{n_{train}}]^T$ denotes the similarity between the selected global class representation $\xi_i$ and the query sample $x_k$.

By combining the registration loss and classification loss of query images together, the total loss function for a training iteration is given as in Equation \ref{total_loss}, and the outline of computing training episode loss is given in Algorithm \ref{alg:fsl}. The loss will update all learnable components including global representations, and the parameters of the registration module and feature extractor.
\begin{equation}
\label{total_loss}
\small
\begin{split}
&\mathcal{L}_{total}(S,Q)=\sum_{c_i\in S}\mathcal{L}_{reg}(r_{c_i})+\sum_{k\in Q}\mathcal{L}_{fsl}(x_k)
\end{split}	
\end{equation}

During the test, we use the same procedure to predict the label of unlabeled data. That is, we first feed support set to feature extractor and get episodic class representations for each class. Then, episodic class representations are used to register the corresponding global class representation via the registration module. After that, we perform the nearest neighbor search by computing the Euclidean distance between the feature vector of a test sample and the selected global representations. 

\subsection{Extension to Generalized FSL}
Although the proposed approach is originally designed for standard FSL, it can be easily extended to generalized FSL: simply including test data from both base and novel classes, and their labels are predicted from all $N$ classes in $C_{total}$ in the test stage. This setting is much more challenging and realistic than the standard FSL, where test data are from only novel classes. Note that, our registration module inherently is a classifier for generalized FSL setting. Our registration module not only optimizes novel class representations but updates base class representations as well. By comparing each test sample against global representations of both base classes and novel classes, our registration module can directly predict the probability of the test images belong to each class in $C_{total}$. 


\section{Experiments and Discussion}

In this section, we evaluate our approach by conducting three groups of experiments: 1) standard FSL setting where the label space of test data is restricted to a few novel classes at each test iteration, 2) generalized FSL setting where the label space of test data is extended to both base classes and novel classes, and 3) ablation study.

\begin{table}[t]
\vspace{0.05in}
\begin{center}
\tabcolsep0.07cm
\begin{small}
\begin{tabular}{lcccc}
\specialrule{0.05em}{0pt}{5pt}
\multirow{2}{*}{\bf Model}& \multicolumn{2}{c}{\bf 5 way Acc.}&\multicolumn{2}{c}{\bf 20 way Acc. }\\
&  1 shot &  5 shot & 1 shot &  5 shot \\\specialrule{0.05em}{1pt}{3pt}
MN \cite{vinyals2016bnips}& 97.9 &98.7&93.5&98.7\\
APL \cite{Ramalho2019iclr}&97.9&\textbf{99.9}&97.2&97.6\\
DLM \cite{Triantafillou2017bnips} & 98.8&95.4&99.6&98.6\\
PN \cite{Snell2017nips} &98.8&99.7&96.0&98.9\\
MA \cite{Finn2017icml} & 98.7$\pm$0.4 & \textbf{99.9}$\pm$0.1 & 95.8$\pm$0.3 &98.9$\pm$0.2\\
RN \cite{Sung2018cvpr}& 99.6$\pm$0.2 & 99.8$\pm$0.1 & 97.6$\pm$0.2 &99.1$\pm$0.1\\
MMN \cite{Cai2018cvpr}& 99.28$\pm$0.08 & 99.77$\pm$0.04 & 97.16$\pm$0.10 &98.93$\pm$0.05\\
MG \cite{Zhang2018nips} &99.67$\pm$0.18 & 99.86$\pm$0.11 & 97.64$\pm$0.17 &99.21$\pm$0.10\\
\specialrule{0.05em}{2pt}{2pt}
Ours &  \textbf{99.72}$\pm$0.06 & \textbf{99.90}$\pm$0.10 & \textbf{99.63}$\pm$0.09 & \textbf{99.32}$\pm$0.04\\\specialrule{0.05em}{2pt}{0pt}
\end{tabular}
\end{small}
\end{center}
\vspace{-0.05in}
\caption{Comparative results for FSL on the Omniglot dataset. The averaged accuracy (\%) on 1,000 test episodes is given followed by the standard deviation (\%).}
\label{fsl_Omniglot}
\end{table}

\subsection{Standard Few-Shot Learning}

\subsubsection{Datasets and Settings}

Under the standard FSL setting adopted by all FSL works so far, we evaluate our approach on the most popular benchmarks: Omniglot and miniImageNet. Omniglot \cite{Lake2011ogsci} contains 32,460 images of hand-written characters. It is composed of 1,623 different characters within 50 alphabets. Each character has 20 images. We follow the most common split in \cite{vinyals2016bnips,Snell2017nips}, taking 1,200 characters for training and the rest 423 for testing. Moreover, we adopt the same data preprocessing as in \cite{vinyals2016bnips}: each image is resized to 28 $\times$ 28 pixels and rotated by multiples of 90 degrees as data augmentation. The miniImageNet dataset is a recent collection of ImageNet for FSL. It consists of 100 classes randomly selected from ImageNet \cite{Russakovsky2015ImageNet} and each class contains 600 images with the size of 84 $\times$ 84 pixels. Following the widely used setting in prior works \cite{vinyals2016bnips,Snell2017nips}, we take 64 classes for training, 16 for validation and 20 for testing, respectively. During the training stage, the 64 training classes and 16 validation classes are respectively regarded as base classes and novel classes to decide the hyperparameters of our approach. The reported performance is obtained by our approach trained with 64 training classes as base classes and 20 test classes as novel classes. 

\subsubsection{Implementation Details}

\textbf{Network architecture}: Our feature extractor $F$ mirrors the architecture used by \cite{Snell2017nips,vinyals2016bnips} and consists of four convolutional blocks. Each block comprises a 64-filter 3 $\times$ 3 convolution, batch normalization layer \cite{batchnorm}, a ReLU nonlinearity and a 2 $\times$ 2 max-pooling layer. When applied to the 28 $\times$ 28 Omniglot images, this architecture results in 64-dimensional output space. When applied to the 84 $\times$ 84 miniImageNet images, this architecture results in 1600-dimensional output space. We use the same feature extractor on images in both the support set and query set. The two embeddings $\theta$ and $\phi$ in our registration module use the same architecture: a fully-connected layer followed by a batch normalization layer and a ReLU non-linearity layer. The output channels of the fully-connected layer are 512.

\noindent\textbf{Training procedure}: We first train our feature extractor $F$ for simple classification task by using all base class. Each global class representation is then initialized by first using the pretrained $F$ to extract visual features of images from its class and then averaging these visual features. The data hallucinator \cite{Wang2018cvprlsl} used in Section \ref{sec:syn} is pretrained with the pretrained $F$ as feature extractor. The registration module is trained from scratch with random Gaussian initialization. After initializing the feature extractor, global representations, the data hallucinator, and the registration module, we train them together in an end-to-end manner. Stochastic gradient descent (SGD) \cite{sgd} with momentum is used for model training with a base learning rate of 0.001 and a momentum of 0.9. The learning rate is annealed by 1/10 for every 3,000 episodes. 

\begin{table}[t]
\begin{center}
\tabcolsep0.3cm
\begin{small}
\begin{tabular}{lcc}
\specialrule{0.05em}{0pt}{3pt}
\multirow{2}{*}{\bf Model}& \multicolumn{2}{c}{\bf 5 way Acc.}\\
&  1 shot & 5 shot \\\specialrule{0.05em}{2pt}{2pt}
MLSTM \cite{Ravi2017iclr}& 43.44 $\pm$ 0.77 &60.60 $\pm$ 0.71\\
MN \cite{vinyals2016bnips}& 43.56 $\pm$ 0.84 &55.31 $\pm$ 0.73\\
MA \cite{Finn2017icml}& 48.70 $\pm$ 1.84 & 63.11 $\pm$ 0.92\\
PN \cite{Snell2017nips} &49.42 $\pm$ 0.78 &68.20 $\pm$ 0.66 \\
DLM \cite{Triantafillou2017bnips} &50.28 $\pm$ 0.80 & 63.70 $\pm$ 0.70\\
RN \cite{Snell2017nips} &50.44 $\pm$ 0.82 &65.32 $\pm$ 0.70 \\
MG \cite{Zhang2018nips} &52.71 $\pm$ 0.64 & 68.63 $\pm$ 0.67\\
MMN \cite{Cai2018cvpr}& \textbf{53.37} $\pm$ 0.48 & 66.97 $\pm$ 0.35 \\
\specialrule{0.05em}{2pt}{2pt}
Ours & 53.21 $\pm$ 0.40&\textbf{72.34} $\pm$ 0.32\\
\specialrule{0.05em}{2pt}{0pt}
\end{tabular}
\end{small}
\end{center}
\vspace{-0.0in}
\caption{Comparative results for FSL on the miniImageNet dataset. The averaged accuracy (\%) on 600 test episodes is given followed by the standard deviation (\%).}
\label{fsl_miniImageNet}
\end{table}

\subsubsection{Results on Omniglot}

Following the standard setting adopted by most existing few-shot learning works, we conduct 5-way 1-shot/5-shot and 20-way 1-shot/5-shot classification on the Omniglot dataset. In the four FSL tasks, each training episode contains 60 classes and each test episode contains $n_{test}$ classes ($n_{test} = 20$ for 20-way scenario and $n_{test} = 5$ for 5-way scenario). In 1-shot and 5-shot scenarios, each query set has 5 images per class, while each support set contains 1 and 5 image(s) per class, respectively. For a training episode, images in the support sets and query sets are randomly selected from the whole training set. In a test episode, images in the support sets are randomly selected from the training set, while images in the query sets are randomly selected from the test set. The evaluation metric is defined as the classification accuracies on randomly selected 1000 test episodes. The comparative results on the Omniglot dataset are provided in Table \ref{fsl_Omniglot}. It can be observed that our approach has achieved a new state-of-the-art performance. This validates the effectiveness of our approach due to its unique global class representation learning strategy. 
\vspace{-0.05in}

\subsubsection{Results on miniImageNet}
\label{sec: mini}
Following previous works \cite{vinyals2016bnips,Snell2017nips}, we conducted 5-way 1-shot and 5-way 5-shot classification on the miniImageNet dataset. The 5-way 1-shot and 5-way 5-shot on the miniImageNet dataset is similar to that on the Omniglot dataset, except three differences: 1) In 5-way 1-shot FSL, each training episode contains 30 classes; 2) In 5-way 5-shot FSL, each training episode contains 20 classes; 3) Each query set has 5 images per class in training and test episodes. The evaluation metric is defined as the classification accuracy on randomly selected 600 test episodes. Table \ref{fsl_miniImageNet} provides comparative results for FSL on the miniImageNet dataset. We can see that our approach significantly outperforms other FSL alternatives on 5-way 5-shot setting and achieves the joint best results under 5-way 1-shot setting. Our registration module yields 100\% registration accuracy on the test data and the similarity scores are close to their one-hot labels. This indicates that the registration module can accurately select the corresponding global representations for episodic class representations of support sets. That is, the soft manner of registering proposed in Section \ref{sec:FSL} has achieved the same performance as the `argmax' operation on similarity score.


\subsection{Generalized Few-Shot Learning}
\label{sec: gfsl}
\subsubsection{Dataset and Settings}

To further evaluate the effectiveness of our approach, we test our approach in a more challenging yet practical setting, i.e., generalized FSL, where the label space of test data is extended to both base and novel classes. We conduct experiments on 5-way 5-shot FSL on the miniImageNet dataset with a new data split. Concretely, we use the same class split as the original miniImageNet (i.e., training/validation/test: 64/16/20), with a new sample split: we randomly select 500 images of the total of 600 images per base class and a few samples per novel class to form a new training set. We select 100 images per base/novel class from the remaining data to form a new test set. The hyperparameter selection strategy is the same as that in standard FSL.

Inspired by generalized ZSL \cite{Changpinyo2016CVPR,Changpinyo2017ICCV,Kodirov2017CVPR,Romera2015icml,chao2016empirical}, we define three evaluation metrics for generalized FSL : 1) $acc_a$ -- the accuracy of classifying the all test samples to all the classes. 2) $acc_b$ -- the accuracy of classifying the data samples from the base classes to all the classes (both base and novel). 3) $accu_n$ -- the accuracy of classifying the data samples from the novel classes to all the classes. Note that, test examples are from both the base and novel classes and each approach has to predict labels from the joint label space.

\begin{table}[t]
\vspace{0.0in}
\begin{center}
\tabcolsep0.3cm
\begin{small}
\begin{tabular}{lcccc}
\specialrule{0.05em}{0pt}{3pt}
{\bf Model}& {\bf $accu_a$}&{\bf $accu_b$}&{\bf $accu_n$}
\\\specialrule{0.05em}{2pt}{2pt}
MN \cite{vinyals2016bnips}&26.98&33.54&0.75\\
PN \cite{Snell2017nips} &31.17&39.53&0.52\\
RN \cite{Sung2018cvpr}&32.48&40.24&1.42\\
\specialrule{0.05em}{2pt}{2pt}
Ours &\textbf{39.14}&\textbf{46.32}&\textbf{12.98}\\
\specialrule{0.05em}{2pt}{0pt}
\end{tabular}
\end{small}
\end{center}
\vspace{-0.01in}
\caption{Comparative results (\%) on the miniImageNet dataset under the generalized FSL setting. In this setting, test examples are from both the base and novel classes and each approach has to predict labels from the joint label space. Notations: $acc_a$ -- the accuracy of classifying the all test samples to all the classes (both base and novel). $acc_b$ -- the accuracy of classifying the data samples from the base classes to all the classes. $accu_n$ -- the accuracy of classifying the data samples from the novel classes to all the classes.}\vspace{-0.2in}
\label{gfsl_miniImageNet}
\end{table}

We compare our model with three recent approaches\footnote{The results of these three approaches are obtained by training the original code provided in their papers using our new split of miniImageNet.}: 1) PN \cite{Snell2017nips} which recognizes unlabeled images based on distances from each class mean in a learned embedding space. 2) MN \cite{vinyals2016bnips} which recognizes unlabeled data by a soft nearest neighbor mechanism with the outputs of a contextual embedding as references. The contextual embedding is trained with images from support sets and query sets to emphasize features that are relevant for the particular query class. 3) RN \cite{Sung2018cvpr} which recognizes unlabeled images by using a Relation Network learned with training set to compute relation scores between query images and class mean. These three methods can be easily extended to generalized FSL. Concretely, the model training is almost the same as standard FSL setting, except that new data split is used and few-shot samples from novel classes are included in the training set. During the test stage, we formulate the generalized FSL as a 100-way FSL problem, and all test data are classified into the joint space of both base and novel classes. In \cite{Snell2017nips}, we average features of all samples within a class as a class mean to recognize test data. Similarly, features of test samples and average features of all samples within each class are feed into the Relation Network and Context embedding in \cite{Sung2018cvpr} and \cite{vinyals2016bnips}, respectively.  

\subsubsection{Results}

Table \ref{gfsl_miniImageNet} provides the comparative results of generalized FSL on the miniImageNet dataset. We can observe that: 1) Our approach achieves the best results on all evaluation metrics, with bigger margins than those under the standard setting. This shows that our model has the strongest generalization ability under this more challenging setting. 2) Our approach outperforms the PN and RN, because we learn global class representation for each class, while they estimate episodic class representations. 3) MN yields much lower results than our approaches. It is expected: context embedding encodes examples of \emph{all} classes; with so many base class examples, they overwhelm those in the novel classes, making context embedding fail to emphasize novel class features. Our sample synthesis strategy increases intra-class variance and thus alleviates the data scarcity issue in novel classes.

\begin{figure}[t]
\begin{center}
\includegraphics[width=0.9\columnwidth]{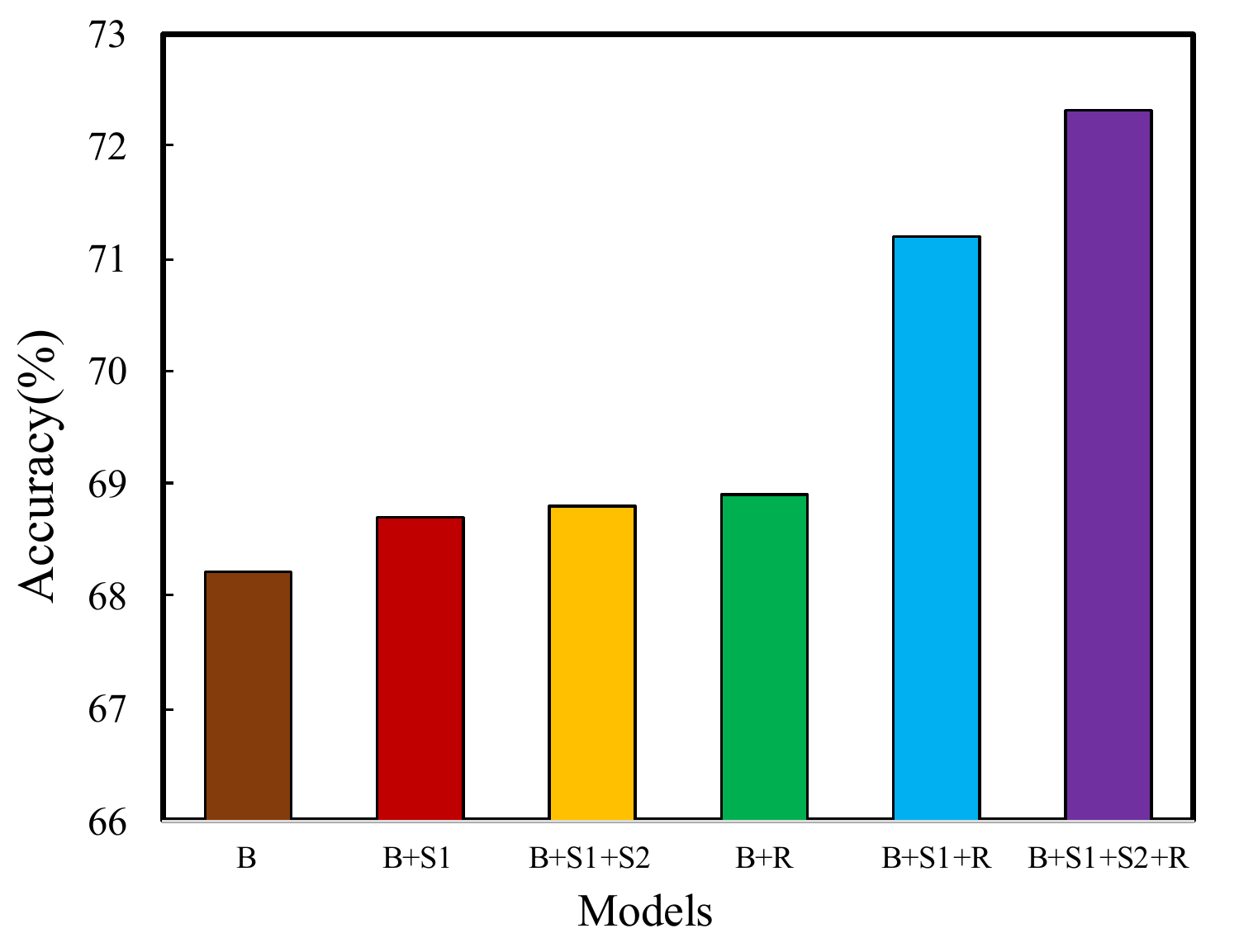}
\end{center}
\vspace{-0.1in}
\caption{Ablation study for FSL on the miniImageNet dataset under the standard FSL setting. Different methods to obtain global class representations are denoted as follows: `B' -- Averaging the visual features in the same class to obtain the class representations; `S1' -- The first step of the method proposed in Section \ref{sec:syn}; `S2' -- The second step of the method proposed in Section \ref{sec:syn}; `R' -- registration module proposed in Section \ref{sec:reg}. }
\label{ab_study}
\vspace{-0.05in}
\end{figure}

\subsection{Ablation Study}
\label{ab}
\subsubsection{Components Analysis}
We compare our full model with a number of stripped-down versions to evaluate the effectiveness of the key components of our approach. Before introducing our methods for ablation study, we denote the key components of our approach as follows: `B' -- Averaging the visual features in the same class to obtain class representations; `S1' -- The first step of the method proposed in Section \ref{sec:syn}; `S2' -- The second step of the method proposed in Section \ref{sec:syn}; `R' -- registration module proposed in Section \ref{sec:reg}. By combining the key components, we compare six FSL models, each of them uses the same FSL framework and differs only in how to learn global class representations: 1) `B'-- The global representations of both base/novel class are obtained by `B'; 2) `B+S1' -- The global representations of base classes are obtained by `B' ; while the global representations of novel classes are obtained by `S1'; 3) `B+S1+S2' -- The global representations of base classes are obtained by `B' ; while global representations of novel classes are obtained by `S1' followed by 'S2'; 4) `B+R' -- The episodic representation of each base/novel class is obtained by 'B', and the global class representations of both base and novel classes are learned by `R' with these episodic representations as inputs;. 5) `B+S1+R' -- The episodic representation of each base class is obtained by 'B', while the episodic representation of each novel class is obtained by 'S1'. The global class representations of both base and novel classes are learned by `R' with these episodic representations as inputs; 6) `B+S1+S2+R' -- Our full model. 

The ablation study results in Figure \ref{ab_study} show that: leveraging our sample synthesis strategy or registration module alone cannot well learn the global representation of classes (see `B+S1' vs. `B', `B+S1+S2' vs. `B' and `B+R' vs. `B'). However, when both of the two methods are used simultaneously to learn global class representations, performance has been significantly improved (see `B+S1+R' vs. `B' and `B+S1+S2+R' vs. `B'). It is expected because: 1) When applying sample synthesis strategy alone, we use an episodic representation as a global representation. Although the synthesis strategy can increase intra-class variance, the episodic representation loses the global class consistency, which limits performance improvement. 2) When applying the registration module alone, the severe class imbalance issue will limit performance improvement. 3) By integrating both of the two methods into the FSL framework, our approach can address the above two issues and the performance thus will significant improve.  These results clearly illustrate the effectiveness of these key components in our approach.

\subsubsection{New Novel Classes}
Our method is flexible to adapt to new unseen novel classes: the model learns the global representations of new unseen novel classes only with its model parameters and the global representations of base and seen novel classes fixed. This means that our method has a low cost for adding new unseen classes. To validate this, we have conducted an additional experiment:  we extend  Mini-ImageNet (whole 100 classes) with another 20 ImageNet classes as the new unseen novel classes. Each new unseen novel class has 5 training samples and 100 test samples. We test our model under the generalized FSL setting described in Section~\ref{sec: gfsl}, where the model has to predict labels from the joint label space of all 120 classes, which include 100 Mini-ImageNet classes and 20 new novel classes. 
Table \ref{tab1} below shows that even when the 100 seen class global representations are fixed, our model is still able to beat Prototypical Net by a large margin. 

\begin{table}[htbp]
\vspace{-0.02in}
\begin{center}
\tabcolsep0.3cm
\begin{small}
\begin{tabular}{lcccc}
\hline
Model& {\bf $accu_a$}&{\bf $accu_b$}&{\bf $accu_n$}
\\\hline
PN \cite{Snell2017nips} &26.80\%&31.62\%&1.07\%\\
Ours &\textbf{30.36\%}&\textbf{40.20\%}&\textbf{12.60\%}\\
\hline
\end{tabular}
\end{small}
\end{center}
\vspace{-0.05in}
\caption{Comparative results under generalized FSL setting on 100 Mini-ImageNet classes and 20 new novel classes. Notations: $accu_a$ -- the accuracy of classifying all test samples to all the classes (both 100 Mini-ImageNet classes and 20 new novel classes). $accu_b$ -- the accuracy of classifying the data samples from the 100 Mini-ImageNet classes to all the classes. $accu_n$ -- the accuracy of classifying the data sampled from the 20 new novel classes to all the classes.}\vspace{-0.03in}
\label{tab1}
\end{table}

\section{Conclusion}
We proposed to solve the challenging FSL problem by learning a global class representation using both base and novel class training samples. In each training episode, an episodic class mean computed from a support set is registered with the global representation via a registration module. This produces a registered global class representation for computing the classification loss using a query set. Our approach can be easily extended to the more challenging generalized FSL setting. Our approach is shown to be effective on both standard FSL and generalized FSL. 

\noindent\textbf{Acknowledgements} This work is supported by National Basic Research Program of China (2015CB352502), NSFC (61573026), and BJNSF (L172037).

{\small
\bibliographystyle{ieee_fullname}
\bibliography{ktsi_bib}
}

\end{document}